\documentclass[10pt,twocolumn,letterpaper]{article}

\usepackage[pagenumbers]{cvpr} 

\usepackage{multirow}
\usepackage{makecell}
\usepackage{pifont}
\usepackage{algorithm}
\usepackage{algpseudocode}
\algrenewcommand\algorithmicindent{1.0em}  

\definecolor{cvprblue}{rgb}{0.21,0.49,0.74}
\usepackage[pagebackref,breaklinks,colorlinks,allcolors=cvprblue]{hyperref}

\def\paperID{4339} 
\def\confName{CVPR}
\def\confYear{2026}

\title{Rethinking Two-Stage Referring-by-Tracking in Referring Multi-Object Tracking: Make it Strong Again}

\author{Weize Li\\ \and Yunhao Du\\ \and Qixiang Yin\\ \and Zhicheng Zhao\\ \and Fei Su
}

\begin{document}
\maketitle
\begin{abstract}

Referring Multi-Object Tracking (RMOT) aims to track multiple objects specified by natural language expressions in videos. 
With the recent significant progress of one-stage methods, the two-stage Referring-by-Tracking (RBT) paradigm has gradually lost its popularity. However, its lower training cost and flexible incremental deployment remain irreplaceable. Rethinking existing two-stage RBT frameworks, we identify two fundamental limitations: the overly heuristic feature construction and fragile correspondence modeling. To address these issues, we propose FlexHook, a novel two-stage RBT framework. In FlexHook, the proposed Conditioning Hook (C-Hook) redefines the feature construction by a sampling-based strategy and language-conditioned cue injection. Then, we introduce a Pairwise Correspondence Decoder (PCD) that replaces CLIP-based similarity matching with active correspondence modeling, yielding a more flexible and robust strategy. Extensive experiments on multiple benchmarks (Refer-KITTI/v2, Refer-Dance, and LaMOT) demonstrate that FlexHook becomes the first two-stage RBT approach to comprehensively outperform current state-of-the-art methods.
Code can be found in \url{https://github.com/buptLwz/FlexHook}

\end{abstract}    
\section{Introduction}
\label{sec:intro}

By integrating natural language processing into traditional Multi-Object Tracking (MOT), Referring Multi-Object Tracking (RMOT)~\cite{Wu_2023_CVPR} aims to achieve consistent tracking of specific targets through precise linguistic guidance. Existing RMOT approaches can be broadly divided into three paradigms:

\begin{figure}[tb]
    \centering
    \includegraphics[width=0.9\linewidth]{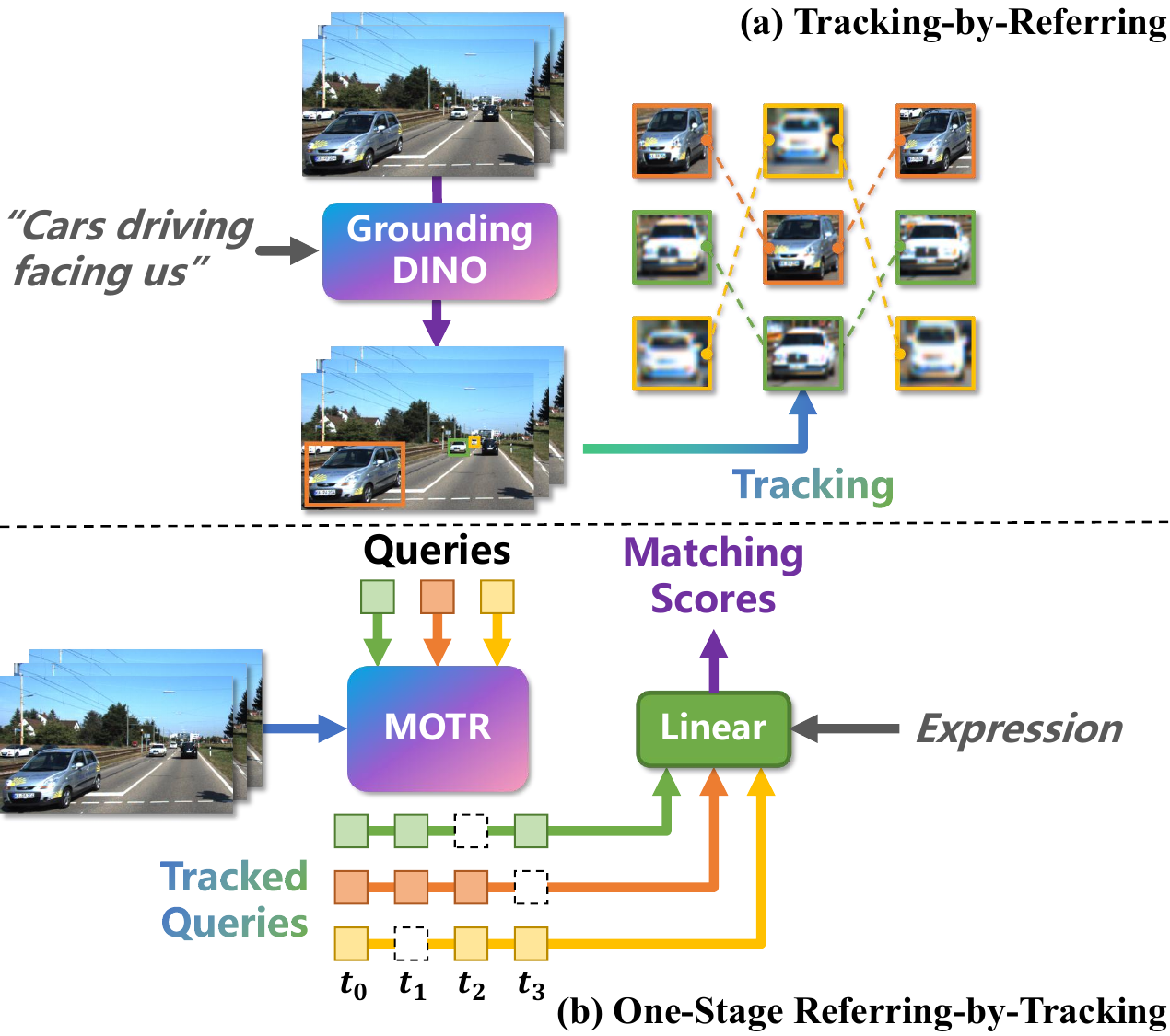}
    \caption{(a) Tracking-by-Referring associates trajectories within boxes located by GroundingDINO~\cite{liu2023grounding}. (b) One-stage Referring-by-Tracking projects queries decoded by MOTR~\cite{10.1007/978-3-031-19812-0_38} to matching scores based on expressions. }
    \label{fig:vs}
\end{figure}

(1) Tracking-by-Referring (TBR), a newly emerging paradigm represented by LaMOTer~\cite{li2024lamotlanguageguidedmultiobjecttracking}, emphasizes the referring process and constructs trajectories by associating detection boxes predicted by GroundingDINO~\cite{liu2023grounding}, as shown in Fig.~\ref{fig:vs} (a).
While benefiting from large vision-language models (VLMs) for open-set generalization, its development remains in an early stage.

(2) One-stage Referring-by-Tracking (RBT), introduced by TransRMOT~\cite{Wu_2023_CVPR}, has advanced rapidly in recent years~\cite{zhang2024bootstrappingreferringmultiobjecttracking,ma2024mlstrackmultilevelsemanticinteraction,li2025visual,zhuang2025cgatracker,Zhao_Hao_Zhang_Liu_Li_Sui_He_Chen_2025}. These methods directly compute matching scores using trajectory queries decoded by MOTR~\cite{10.1007/978-3-031-19812-0_38}, requiring an end-to-end joint optimization of detection, tracking, and referring as shown in Fig.~\ref{fig:vs} (b).

\begin{figure*}[ht]
    \centering
    \includegraphics[width=0.9\linewidth]{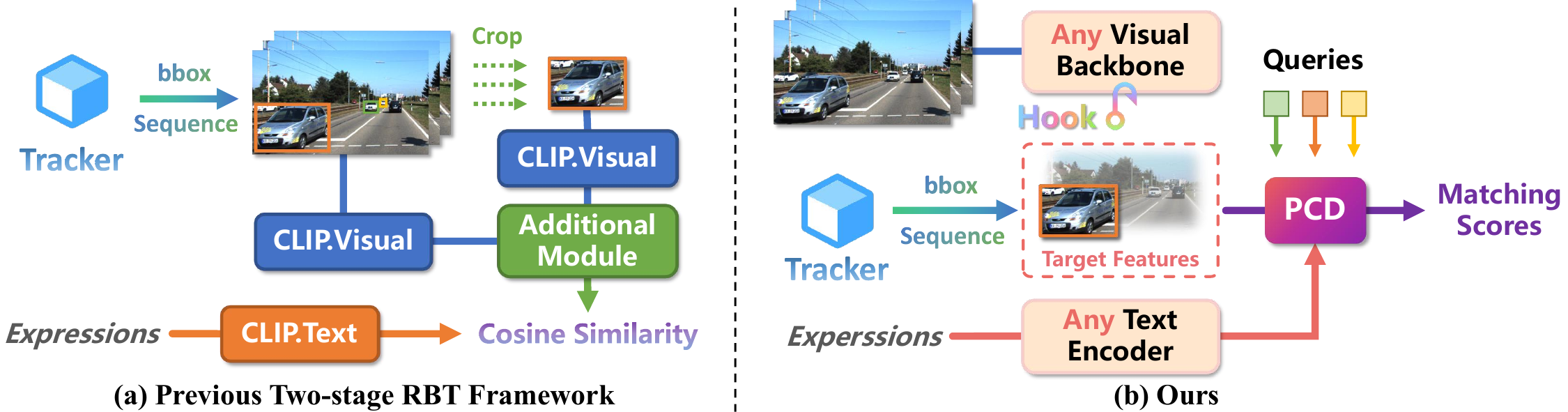}
\caption{(a) Previous two-stage RBT reuses the encoder and computes CLIP-based~\cite{pmlr-v139-radford21a} cosine similarity as the matching score.
(b) Our method directly hooks features from the visual backbone and decodes scores via PCD in a composed feature space.}
\label{fig:core}
\end{figure*}

(3) Two-stage RBT, proposed by iKUN~\cite{Du_2024_CVPR}, fully decouples the tracking and referring processes, as shown in Fig.~\ref{fig:core} (a). It leverages off-the-shelf trackers to perform tracking and directly constructs visual features from the tracked boxes, significantly reducing the burden of joint optimization and enabling flexible, incremental extension to deployed tracking systems or future advances. 




However, despite being introduced around the same period, the development of two-stage RBT has lagged far behind its one-stage counterpart. In particular, iKUN~\cite{Du_2024_CVPR} achieves only 10.32 HOTA on the Refer-KITTI-v2~\cite{zhang2024bootstrappingreferringmultiobjecttracking} dataset, highlighting the limitations of current two-stage RBT frameworks.
Nevertheless, among existing methods, the advantages of the two-stage RBT paradigm in training efficiency and incremental deployment remain irreplaceable. Motivated by this observation, we rethink the design rationality of the current two-stage RBT paradigm and identify two fundamental issues that hinder its full potential:

(1) \textbf{Overly Heuristic Feature Construction}. Two-stage RBT typically employs a visual backbone to reconstruct target features from frames and tracking results, avoiding access to preceding subtasks and preserving incremental flexibility, as shown in Fig.~\ref{fig:core} (a). To further capture spatial relationships required by RMOT, existing methods often dual-encode both the full image and cropped target patches using a shared encoder along with additional fusion modules.

Although this approach may appear intuitive, on one hand, it overlooks an important property of modern visual backbones (e.g. ResNet~\cite{He_2016_CVPR}, Swin~\cite{Liu_2021_ICCV}), which inherently provide continuous spatial gradient flow and strong context aggregation after pretraining. 
Current designs unnecessarily duplicate the backbone computation, forcing the model to relearn semantic relationships through retrained modules rather than leveraging the pretrained capabilities.

On the other hand, different types of contextual expressions (e.g., positional or directional) require adaptive focus on varying regions, yet existing language-agnostic feature modeling fails to regulate attention accordingly.

(2) \textbf{Fragile Correspondence Modeling}. As shown in Fig.~\ref{fig:core} (a), two-stage RBT methods primarily establish target–expression correspondence by computing cosine similarity between visual and textual embeddings, which relies heavily on CLIP’s~\cite{pmlr-v139-radford21a} pretrained alignment space. However, RMOT’s requirement for exhaustive target awareness and contextual cues unavoidably demands learning knowledge beyond CLIP’s~\cite{pmlr-v139-radford21a} pretraining scope. Once additional modules are introduced, as discussed above, or when the visual backbone is replaced with a dense alternative, the original CLIP-based~\cite{pmlr-v139-radford21a} similarity measure easily breaks down. This correspondence modeling strategy actually imposes an upper bound on achievable alignment and restricts the framework’s scalability and long-term generalization.

To overcome these limitations, we propose FlexHook, a novel two-stage RBT framework as shown in Fig.~\ref{fig:core} (b). 
FlexHook operates like a hook function in programming~\cite{lopez2017survey}, redefining the feature construction paradigm while fully preserving the original backbone flow and enabling seamless integration with any visual/text encoder.
Specifically, we propose a Conditioning Hook (C-Hook) module that directly samples, rather than models, target features from the visual backbone to restore the contextual gradient flow, and employs language-conditioned auxiliary sampling points to inject conditional cues.
In addition, we design a Pairwise Correspondence Decoder (PCD) that performs learnable pairwise discrimination in place of the static cosine-similarity metric. PCD converts passive similarity comparison into active correspondence modeling at the pairwise level, thus removing dependence on CLIP~\cite{pmlr-v139-radford21a}. Our contributions can be summarized as follows:

(1) The proposed C-Hook redefines feature construction in two-stage RBT by a sampling-based strategy, restoring original contextual gradient flow of visual backbone.

(2) We incorporate language-conditioned auxiliary sampling points to extract language-aware cues, enabling adaptive disentanglement of complex referring expressions.

(3) We propose the PCD, replacing static cosine similarity with learnable pairwise discrimination, converting passive similarity matching into active correspondence modeling and eliminates dependence on CLIP.

(4) FlexHook revolutionizes two-stage RBT paradigm and makes it strong again. It is the first two-stage RBT framework that significantly outperforms existing state-of-the-art methods by a large margin.

\section{Related Work}

\begin{figure*}[t]
    \centering
    \includegraphics[width=0.95\linewidth]{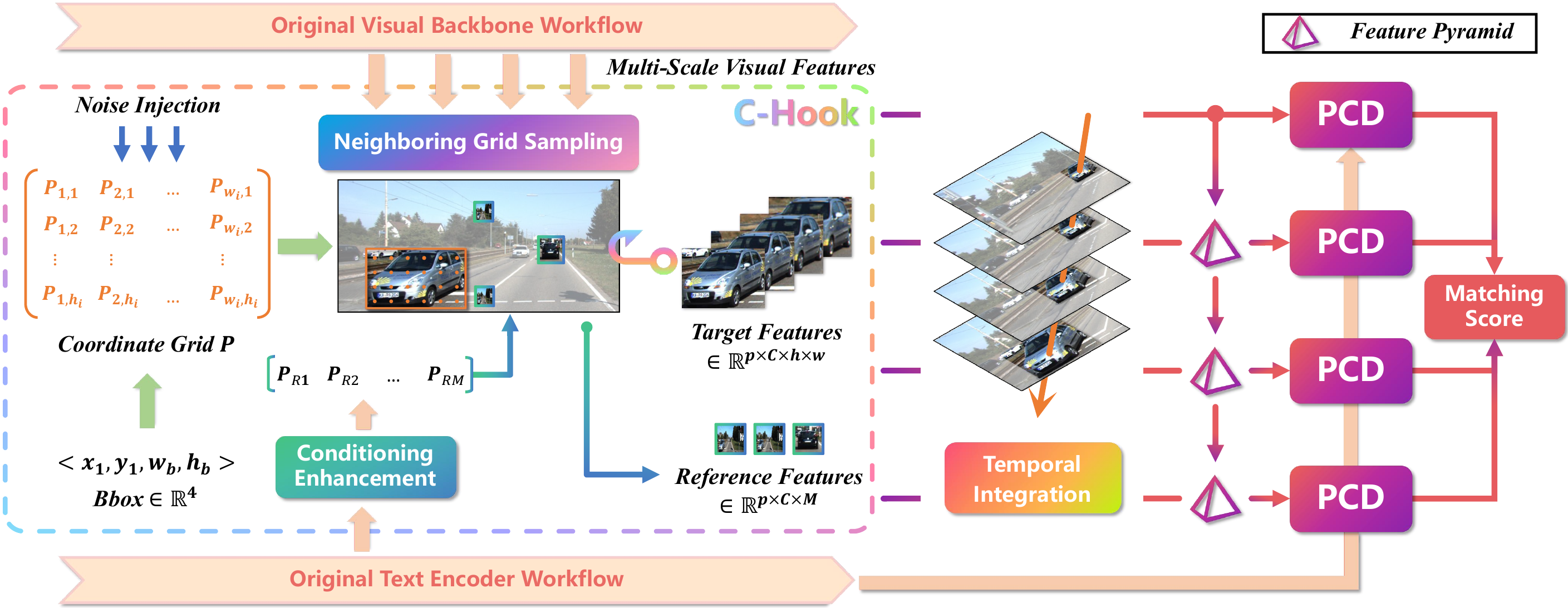}
\caption{The overall framework of FlexHook. FlexHook directly extracts features from multi-scale feature maps via C-Hook during the original workflow, without any additional encoding stages. C-Hook consists of two components: Neighboring Grid Sampling, which samples target features $F_J$ based on trajectory bounding boxes $\mathcal{B}^i_{t:t+p}$, and Conditioning Enhancement, which samples reference features $F_r$ conditioned on linguistic features $F_l$. The sampled features are fused across frames $p$ through Temporal Integration. The multi-scale features are finally aggregated using a feature pyramid~\cite{Lin_2017_CVPR} and decoded layer-by-layer in PCD to generate the final matching scores.}
    \label{fig:overall}
\end{figure*}
\subsection{Referring Single Object Tracking}


Referring Single Object Tracking (RSOT) has been studied for several years. Most SOTA methods~\cite{Botach_2022_CVPR,Wu_2022_CVPR,10.1145/3664647.3680657,10.1145/3726529} follow classic tracking strategies, integrating advancements in referring video segmentation~\cite{Ding_2023_ICCV} (or detection~\cite{WEN2020102907}) and focusing on end-to-end joint tracking algorithms. Previous methods~\cite{ZHAO202310,Zhou_2023_CVPR,10208210,kang2025exploring,10149530} often used initial bounding boxes, templates, or zero-padding as input for the first frame, and then continuously tracked a single target in subsequent frames through matching algorithms. Recently, MENDER~\cite{NEURIPS2023_098491b3} proposed the GroOT dataset and, via third-order tensor decomposition, enabled tracking a single target with natural language. Although significant progress has been made~\cite{10633272,Wu_Han_Liu_Wang_Xu_Zhang_Shen_2025}, RSOT requires a precise target reference to ensure the tracking of a single object, which differs substantially from textual prompts in real-world scenarios and poses significant annotation challenges.


\subsection{Referring Multi Object Tracking}

\label{sec:formatting}
TransRMOT~\cite{Wu_2023_CVPR} pioneered RMOT by introducing the Refer-KITTI dataset along with a baseline model based on MOTR~\cite{10.1007/978-3-031-19812-0_38}. To address the high computational cost of these methods, iKUN~\cite{Du_2024_CVPR} innovatively proposed the two-stage RBT paradigm. Then, TempRMOT~\cite{zhang2024bootstrappingreferringmultiobjecttracking} constructs the enhanced Refer-KITTI-V2 dataset by leveraging large language models~\cite{NEURIPS2020_1457c0d6}. However, one-stage RBT methods built upon TransRMOT have been gradually gaining popularity by introducing novel temporal modeling and decoding mechanisms~\cite{ma2024mlstrackmultilevelsemanticinteraction,li2025visual,zhuang2025cgatracker,Zhao_Hao_Zhang_Liu_Li_Sui_He_Chen_2025}. Recently, focusing on RMOT performance under open-vocabulary scenarios, LaMOTer~\cite{li2024lamotlanguageguidedmultiobjecttracking} introduces a Tracking-by-Referring architecture to further enrich the logic of RMOT. Despite recent progress, the practical advantages of two-stage RBT pipelines in incremental extension and training efficiency remain irreplaceable, but have received little attention.

\section{Method}

\subsection{Problem Formulation}

In RMOT, the trajectory of target $i$ can be represented by a sequence of detection boxes across frames, denoted $\mathcal{B}^i = \{B^i_t\}_{t = t^i}^{T^i}$, where $t$ indicates the frame index, and $t^i$ and $T^i$ denote the starting and ending frames in which the target appears. We use $\mathcal{B}^i_{a:b}$ to denote a segment of target $i$'s trajectory, where $a$ and $b$ represent the starting and ending frames of this segment, respectively. The index $t$ is not necessarily continuous, as the target may disappear or be occluded in certain frames and reappear later. For each video, let $\mathcal{O} = \{\mathcal{B}^i\}_{i=1}^{K}$ be the set of all $K$ target trajectories, and let $\mathcal{E} = \{E_j\}_{j=1}^{N}$ denote the set of $N$ natural language expressions annotated for the targets in the whole video. The process of inferring all target trajectories $\mathcal{O}$ from the video frame sequence $\mathcal{I} = \{I_t\}_{t=0}^T$ constitutes the tracking subtask, while the referring subtask can be formulated as learning a many-to-many matching relation between the expression set $\mathcal{E}$ and the trajectory set $\mathcal{O}$: 
\begin{equation}
    \mathcal{R} \subseteq \mathcal{E} \times \mathcal{O}
\end{equation}
where $(E_j, \mathcal{B}^i_{x:y}) \in \mathcal{R}$ indicates that expression $E_j$ refers to trajectory segment $\mathcal{B}^i_{x:y}$. Note that the reason we use a segment instead of the entire trajectory is that certain expressions, such as ``people on the left", may become invalid as the target moves. In the two-stage RBT paradigm, no access is available to the detection or tracking stages; instead, the referring subtask is executed directly based on $\mathcal{O}$.

\subsection{Overview}

To facilitate understanding, as illustrated in Fig.~\ref{fig:overall}, we decompose FlexHook into three main components:
(1) C-Hook, a Conditioning Hook module that constructs context-rich target features and language-conditioned reference features;
(2) Temporal Integration, which integrates visual features across the frame segment to extract temporal motion cues; and
(3) PCD, a Pairwise Correspondence Decoder that produces the matching score for each language–visual feature pair. Given a $p$-frame trajectory segment $\mathcal{B}^i_{t:t+p}$, FlexHook extracts visual features $F_v \in \mathbb{R}^{p  \times H \times W \times C}$ from the corresponding $p$ image frames $\mathcal{I}_{t:t+p}$. Notably, instead of dual-encoding the full image and cropped patches, FlexHook encodes the global image only once. Then, we sample $\hat{N}$ expressions $\hat{\mathcal{E}} = \{E_j\}_{j=1}^{\hat{N}}$ from the expression set $\mathcal{E}$ and extract linguistic features $F_l \in \mathbb{R}^{\hat{N} \times L \times C}$. 
Based on them, we repeat the workflow of C-Hook sampling, Temporal Integration, and PCD decoding on the feature maps from each layer of the visual backbone, and connect them through a feature pyramid.
Finally, after averaging the multi-scale results, FlexHook outputs $\hat{N}$ matching scores per forward pass, which represent the correspondence between $\mathcal{B}^i_{t:t+p}$ and one expression in $\hat{\mathcal{E}}$.

\subsection{C-Hook}

C-Hook directly samples both the target features and language-conditioned cues from the backbone's raw feature stream.
It is realized with two components: Neighboring Grid Sampling and Conditioning Enhancement. Its fully sampling-based design enables flexible target focus while preserving the backbone’s original gradient flow, and the injection of linguistic priors further allows it to adaptively modulate reference cues.

\subsubsection{Neighboring Grid Sampling}

In practice, we implement the sampling process through grid sampling, a differentiable operation to preserve gradient flow. However, naive grid sampling alone cannot fully mitigate the distributional inconsistency inherent in the two-stage RBT paradigm: the model is typically trained with ground-truth trajectories, whereas inference relies on trajectory candidates produced by an off-the-shelf tracker.

To address this inherent gap, we introduce neighboring grid sampling to enhance robustness through controlled stochasticity. Concretely, we first convert each bounding box $B^i_t=\langle x_0, y_0, w_b, h_b\rangle$ into a coordinate grid $P_t^i\in \mathbb{R}^{ h\times w \times 2}$, where $(x_0, y_0)$ denotes the upper-left corner of the bounding box, $w$ and $h$ represent the width and height of the sampled features, and $w_b$, $h_b$ indicate the width and height of the bounding box. The grid points $P_t^i(x,y)$ are calculated as
\begin{equation}
P_t^i(x,y) = (x_0+(x-1)\Delta_x, y_0+(y-1) \Delta_y),
\end{equation}
with $\Delta_x=(\tfrac{w_b}{w-1},0)$ and $\Delta_y=(0,\tfrac{h_b}{h-1})$ representing the stride between sampling points. Then, we perform three augmentations to simulate realistic tracking noise: (i) randomly sample fragmented trajectory segments to emulate target loss, (ii) inject Gaussian noise to model localization inaccuracies, and (iii) re-combine grid sequences within a batch to mimic identity switches. The augmented gird enables the model to acquire neighborhood awareness across temporal, spatial, and instance dimensions, enhancing the model’s ability to capture broader contextual information.

Finally, for each $\mathcal{B}^i_{t:t+p}$, we obtain a set of coordinate grids $P \in \mathbb{R}^{p  \times h \times w \times 2}$. Each sampling point is mapped to its location on $F_v$ according to the normalized coordinates, and we sample the target features $J \in \mathbb{R}^{p \times h \times w \times C}$ from $F_v$ via bilinear interpolation.

\subsubsection{Conditioning Enhancement}
\begin{figure}[htbp]
    \centering
    \includegraphics[width=0.7\linewidth]{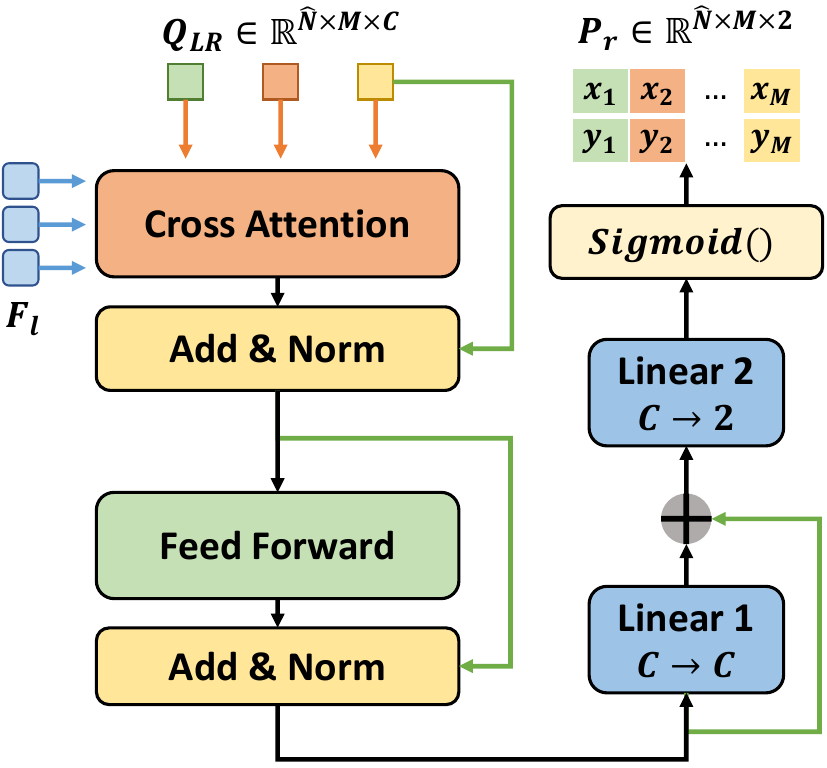}
    \caption{Illustration of Conditioning Enhancement. Guided by the linguistic feature $F_l$, we computes $M$ reference points $P_r$ through a Transformer decoder and a residual MLP followed by a sigmoid function.}
    \label{fig:lre}
\end{figure}

Previous methods typically adopt a language-agnostic setting when constructing target visual features. However, expressions with diverse semantics (e.g., ``the man in red” vs. ``the person on the left”) imply distinct visual focuses and relational cues. To address this discrepancy, we introduce a Conditioning Enhancement module for C-Hook.



To align with the sampling structure of C-Hook, we incorporate linguistic priors by learning additional language-conditioned reference points. 
Specifically, we extract $M$ reference points from each expression, which are decoded from a set of learnable query vectors.
As illustrated in Fig.~\ref{fig:lre}, we perform cross-attention between learnable query vectors $Q_{LR} \in \mathbb{R}^{\hat{N} \times M \times C}$ and the linguistic features $F_l \in \mathbb{R}^{\hat{N} \times L \times C}$ extracted from the text encoder. The attention outputs are then processed by a multi-layer perceptron (MLP) followed by a sigmoid layer to generate normalized 2D reference points $P_r$.
Unlike the coordinate grid $P$, the reference points $P_r$ are time-independent. Therefore, we directly repeat them along the temporal dimension to align with $F_v$, resulting in $P_r \in \mathbb{R}^{p \times \hat{N} \times M \times 2}$. These $P_r$ are used together with the coordinate grid $P$ for grid sampling.

\subsection{Temporal Integration with Optical Flow}

\begin{figure}[htbp]
    \centering
        \includegraphics[width=0.9\linewidth]{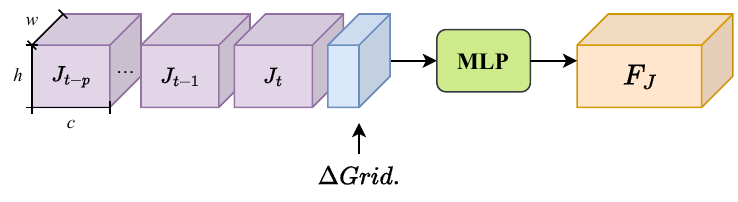}
        \caption{Illustration of the Temporal Integration. We concatenate multi-frame features with grid displacements, then compress them along the channel dimension.}
        \label{fig:ome}
\end{figure}
Motion information is an indispensable part of processing motion expressions (e.g., ``cars turning left") in RMOT. Existing methods often use simple temporal pooling to compress multi-frame information~\cite{Du_2024_CVPR}.
However, given the coordinate grids constructed in C-Hook, it becomes evident that explicit target optical flow can be conveniently derived by computing their differences across frames.
Therefore, as shown in Fig.~\ref{fig:ome}, we concatenate along the channel dimension both the coordinate grid displacements $\Delta Grid= Cat(\{P^i_{t+k}-P^i_{t+k-1}\}_{k=1}^{p})$ and the multi-frame target features $J$ to serve as feature optical flow, producing a fused feature $\in \mathbb{R}^{h\times w \times (p\cdot C+(p-1)\cdot 2)}$. Then, an MLP compresses it to trajectory features $F_J \in \mathbb{R}^{h\times w\times C}$. Since the reference features do not contain motion information of the target, we directly obtain the temporal integrated reference features $F_r \in \mathbb{R}^{\hat{N} \times M \times C}$ through temporal pooling.

\subsection{Pairwise Correspondence Decoder}
To mitigate the limitations induced by fixed CLIP space, we construct a Pairwise Correspondence Decoder (PCD) that directly learns the inherent differences between positive and negative pairs, freeing the framework from the CLIP-based similarity metric.

\begin{figure}[!h]
    \centering
    \includegraphics[width=0.8\linewidth]{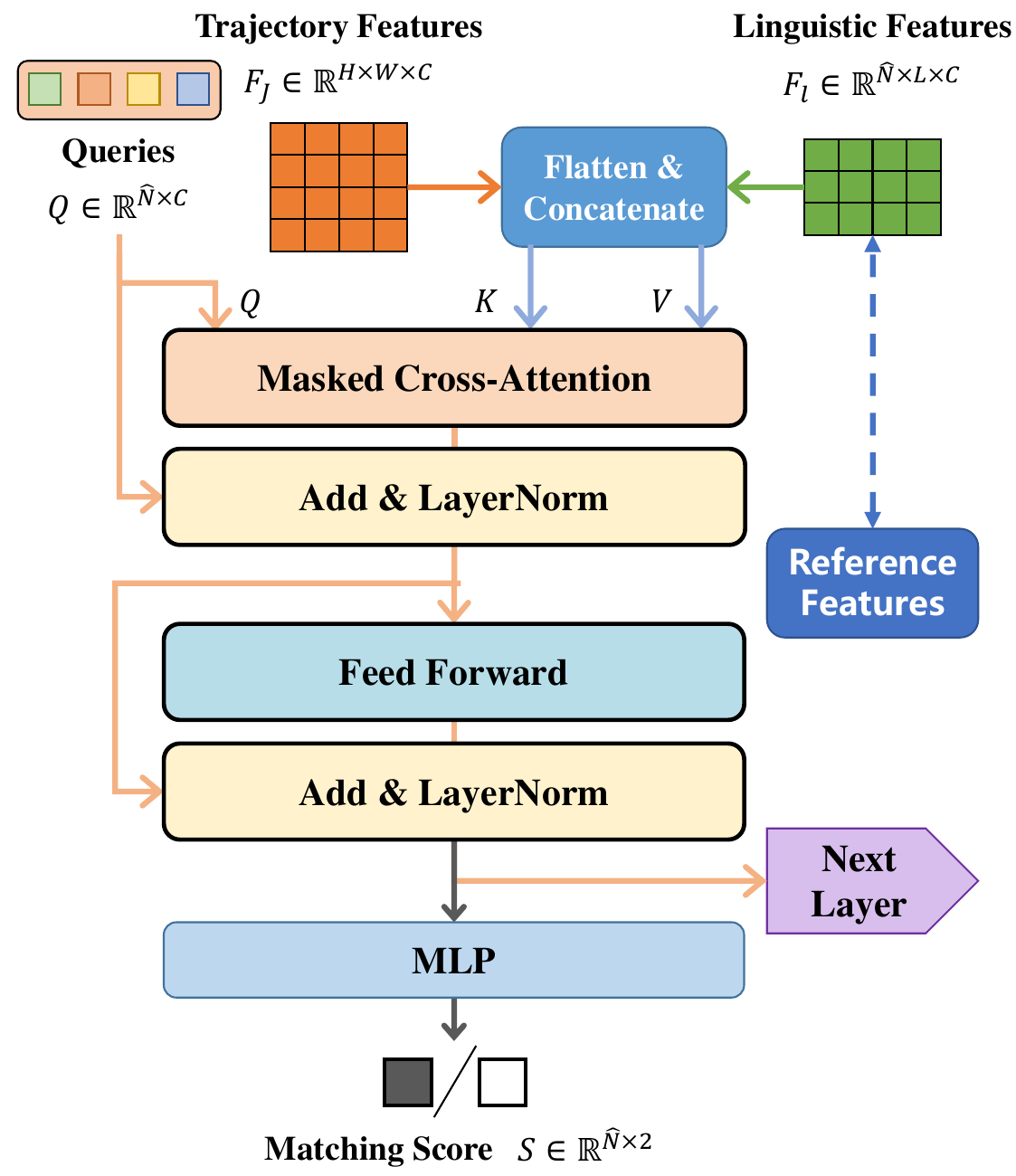}
\caption{Illustration of the PCD. PCD actively examines the matching of sample pairs by performing masked cross-attention between queries and concatenated multi-modal features.}
    \label{fig:pd}
\end{figure}

The $\hat{N}$ sampled expressions and the shared trajectory segment $\mathcal{B}^i_{t:t+p}$ together form $\hat{N}$ sample pairs. Therefore, we use a set of learnable query vectors $Q \in \mathbb{R}^{\hat{N} \times C}$ to extract the matching scores for each pair. Specifically, as shown in Fig.~\ref{fig:pd}, we concatenate the flattened trajectory features $F_J\in\mathbb{R}^{H \cdot W \times C}$, reference features $F_r \in \mathbb{R}^{\hat{N} \cdot M \times C}$, and linguistic features $F_l \in \mathbb{R}^{\hat{N}\cdot L \times C}$ along the first dimension, and then employ them as the Keys and Values, with $Q$ as the Queries, to perform masked cross-attention. By controlling the attention mask $A \in \mathbb{R}^{\hat{N} \times (H \cdot W + \hat{N}\cdot(L+M))}$, all query vectors share the same trajectory features while only accessing their own corresponding linguistic and reference features. This masking strategy ensures pairwise outputs and implicitly enables trajectory features to learn cross-pair knowledge through contrastive learning.


The decoded queries are refined by an FFN and then split into two branches: one feeds an MLP to predict the matching scores $S \in \mathbb{R}^{\hat{N}\times 2}$ at each layer, while the other is passed to the next-layer PCD for further multi-scale decoding.

\subsection{Loss Computation}

We supervise the averaged outputs $\bar S \in \mathbb{R}^{\hat{N}\times2}$ across all layers with the Focal Loss~\cite{Lin_2017_ICCV} to strengthens the model’s multi-scale capability and alleviate the imbalanced sample distribution in the dataset.

In addition, we observe that some learned reference coordinates may collapse towards the boundaries in practice, forming a trivial solution. Therefore, we apply a smooth barrier to penalize points approaching the edges of the normalized space $[-1,1]^2$. Let $Z=(u,v)$ denote a reference point in $P_r$. We define the minimum distance to the boundary $ d_{uv}=\min\big(1-|u|,1-|v|\big)
$, and apply a softplus penalty term to discourage boundary collapse:
\begin{equation}
    \mathcal{L}_{r}
=
\frac{1}{|P_r|}
\sum_{u,v}
\mathrm{softplus}\big(\alpha(\delta-d_{uv})\big),
\end{equation}
where $|P_r|$ denotes the number of points in $P_r$, $\delta$ is a safe margin, and $\alpha$ controls the sharpness. $\mathcal{L}_{r}$ softly pushes the reference points away from the boundary while maintaining stable gradients. The final loss can be formulated as:
\begin{equation}
    \mathcal{L}=\mathcal{L}_{\text{Focal}}(\bar S, S_{\text{gt}})+\lambda\mathcal{L}_r
\end{equation}
where $\lambda$ is used to adjust the penalty strength.

\begin{table*}[t]
\centering
\small
\caption{\textbf{Comparison with state-of-the-art methods} on Refer-KITTI~\cite{Wu_2023_CVPR} and Refer-KITTI-V2~\cite{zhang2024bootstrappingreferringmultiobjecttracking}.}
\begin{tabular}{c|c|c|c|cccc}

\toprule

~&\multicolumn{3}{c|}{Method}& HOTA & DetA & AssA & LocA \\
\midrule

\multicolumn{8}{c}{Refer-KITTI~\cite{Wu_2023_CVPR}} \\
\midrule
\multirow{7}{*}{One-Stage}&\multicolumn{2}{c}{TransRMOT~\cite{Wu_2023_CVPR}}&CVPR'23&46.56&37.97&57.33&90.33 \\
~&\multicolumn{2}{c}{MLS-Track~\cite{ma2024mlstrackmultilevelsemanticinteraction}}&Arxiv'24&49.05&40.03&60.25&- \\
~&\multicolumn{2}{c}{TempRMOT~\cite{zhang2024bootstrappingreferringmultiobjecttracking}}&Arxiv'24&52.21&40.95&66.75&90.40 \\
~&\multicolumn{2}{c}{CGATracker~\cite{zhuang2025cgatracker}}&TCSVT'25&48.81&38.66&61.77&- \\
~&\multicolumn{2}{c}{SKTracker~\cite{li2025visual}}&TMM'25&50.85&42.24&64.39&90.43 \\
~&\multicolumn{2}{c}{HFF-Tracker~\cite{Zhao_Hao_Zhang_Liu_Li_Sui_He_Chen_2025}}&AAAI'25&52.41&41.29&66.65&90.76 \\
~&\multicolumn{2}{c}{DKGTracker~\cite{Li_2025_ICCV}}&ICCV'25&52.08&41.10&66.04&90.54 \\
\midrule

\multirow{4}{*}{Two-Stage}&
\multicolumn{2}{c}{iKUN~\cite{Du_2024_CVPR}}&CVPR'24&48.84&35.74&66.80&- \\

\cmidrule{2-8}
~&\multirow{3}{*}{\textbf{FlexHook}}&CLIP~\cite{pmlr-v139-radford21a}&CLIP-R 50~\cite{pmlr-v139-radford21a}&53.45&42.69&67.00&90.82\\
\cmidrule{3-8}
~&~&\multirow{2}{*}{RoBERTa~\cite{liu2019robertarobustlyoptimizedbert}}&Resnet34~\cite{He_2016_CVPR}&53.60&42.82&\textbf{67.17}&90.82 \\
~&~&~&ROPE Swin-T~\cite{heo2024ropevit}&\textbf{53.83}&\textbf{43.36}&66.92&\textbf{90.87} \\

\midrule
\multicolumn{8}{c}{Refer-KITTI-V2~\cite{zhang2024bootstrappingreferringmultiobjecttracking}} \\
\midrule
\multirow{6}{*}{One-Stage}&\multicolumn{2}{c}{TransRMOT~\cite{Wu_2023_CVPR}}&CVPR'23&31.00&19.40&49.68&89.82 \\
~&\multicolumn{2}{c}{TempRMOT~\cite{zhang2024bootstrappingreferringmultiobjecttracking}}&Arxiv'24&35.04&22.97&53.58&90.07 \\
~&\multicolumn{2}{c}{CGATracker~\cite{zhuang2025cgatracker}}&TCSVT'25&33.19&22.04&50.13&- \\
~&\multicolumn{2}{c}{SKTracker~\cite{li2025visual}}&TMM'25&35.29&23.87&52.35&88.89 \\
~&\multicolumn{2}{c}{HFF-Tracker~\cite{Zhao_Hao_Zhang_Liu_Li_Sui_He_Chen_2025}}&AAAI'25&36.18&24.64&53.27&89.77 \\
~&\multicolumn{2}{c}{DKGTracker~\cite{Li_2025_ICCV}}&ICCV'25&35.26&23.04&54.13&\textbf{91.65} \\

\midrule
~&\multicolumn{2}{c}{iKUN~\cite{Du_2024_CVPR}}&CVPR'24&10.32& 2.17& 49.77& 74.56\\
\cmidrule{2-8}
\multirow{4}{*}{Two-Stage}&\multirow{3}{*}{\textbf{FlexHook}}&CLIP~\cite{pmlr-v139-radford21a}&CLIP-R 50~\cite{pmlr-v139-radford21a}&41.42&29.06&\textbf{59.20}&89.10\\
\cmidrule{3-8}
~&~&\multirow{2}{*}{RoBERTa~\cite{liu2019robertarobustlyoptimizedbert}}&Resnet34~\cite{He_2016_CVPR}&41.50&30.07&57.40&89.31 \\
~&~&~&ROPE Swin-T~\cite{heo2024ropevit}&\textbf{42.53}&\textbf{30.63}&59.19&89.09\\
\cmidrule{2-8}
~&\multicolumn{3}{c|}{\textbf{FlexHook-best} with D-DETR~\cite{zhu2021deformabledetrdeformabletransformers} \& StrongSORT~\cite{du2023strongsort}}&40.73&30.40&54.72&89.39 \\

\bottomrule
\end{tabular} 

\label{tab:sota_refer_kitti_v2}
\end{table*}

\section{Experiment}
\subsection{Experiment Setup}

\subsubsection{Dataset and Metrics} We report the performance of FlexHook on 4 datasets: Refer-Dance~\cite{Du_2024_CVPR}, LaMOT~\cite{li2024lamotlanguageguidedmultiobjecttracking}, Refer-KITTI~\cite{Wu_2023_CVPR}, and Refer-KITTI-V2~\cite{zhang2024bootstrappingreferringmultiobjecttracking}. Among them, Refer-KITTI/V2~\cite{Wu_2023_CVPR,zhang2024bootstrappingreferringmultiobjecttracking} is the most commonly used dataset, focusing solely on autonomous driving scenarios. Refer-Dance~\cite{Du_2024_CVPR} is constructed based on DanceTrack~\cite{Sun_2022_CVPR}. LaMOT~\cite{li2024lamotlanguageguidedmultiobjecttracking} consists of 4 MOT subsets with new natural language annotations, TAO~\cite{10.1007/978-3-030-58558-7_26}, SportsMOT~\cite{Cui_2023_ICCV}, VisDrone~\cite{Wen_2019_ICCV}, and MOT17~\cite{milan2016mot16benchmarkmultiobjecttracking}, making it the largest and most diverse benchmark. However, its TAO~\cite{10.1007/978-3-030-58558-7_26} subset exhibits a strong bias toward evaluating the model’s open-world capability. 

For a fair comparison, we use Higher Order Tracking Accuracy~\cite{luiten2021hota} (HOTA) as the main metric, following TransRMOT~\cite{Wu_2023_CVPR}. HOTA comprehensively considers the detection and association capabilities of the tracking model. We also report the localization accuracy (LocA), the detection accuracy (DetA) and the association accuracy (AssA).
\subsubsection{Detector and Tracker} 
\begin{table}[!ht]
    \centering
    \caption{Details of used detector-tracker. ``D-DETR" denotes the Deformable-DETR~\cite{zhu2021deformabledetrdeformabletransformers}}
    \label{tab:detail}
    \resizebox{0.478\textwidth}{!}{
    \begin{tabular}{c|c|cc}
    \toprule
        \multicolumn{2}{c|}{Dataset} & Detector & Tracker \\ \midrule
        \multirow{4}{*}{LaMOT~\cite{li2024lamotlanguageguidedmultiobjecttracking}} & TAO~\cite{10.1007/978-3-030-58558-7_26} & Co-DETR~\cite{Zong_2023_ICCV} & AED~\cite{11105000} \\ 
        ~ & SportsMOT~\cite{Cui_2023_ICCV} & YOLOX~\cite{yolox2021} & McByte~\cite{Stanczyk_2025_CVPR} \\ 
        ~ & VisDrone~\cite{Wen_2019_ICCV} & YOLOX~\cite{yolox2021} & U2MOT~\cite{liu2023u2mot} \\ 
        ~ & MOT17~\cite{milan2016mot16benchmarkmultiobjecttracking} & YOLOX~\cite{yolox2021} & BoT-SORT~\cite{aharon2022bot} \\ \midrule
        \multicolumn{2}{c|}{Refer-Dance~\cite{Du_2024_CVPR}} & D-DETR~\cite{zhu2021deformabledetrdeformabletransformers} & NeuralSORT~\cite{Du_2024_CVPR} \\ \midrule
        \multicolumn{2}{c|}{Refer-KITTI~\cite{Wu_2023_CVPR}} & TempRMOT-D~\cite{zhang2024bootstrappingreferringmultiobjecttracking} & NeuralSORT~\cite{Du_2024_CVPR} \\ 
        \midrule
        \multicolumn{2}{c|}{\multirowcell{2}{Refer-KITTI-v2~\cite{zhang2024bootstrappingreferringmultiobjecttracking}}} & TempRMOT-D~\cite{zhang2024bootstrappingreferringmultiobjecttracking} & NeuralSORT~\cite{Du_2024_CVPR} \\ 
        \multicolumn{2}{c|}{~}&D-DETR~\cite{zhu2021deformabledetrdeformabletransformers} & StrongSORT~\cite{du2023strongsort}\\ 
        \bottomrule   
        \end{tabular}}
    
\end{table}

\label{sec:detail}

As a two-stage RBT method, FlexHook exhibits strong robustness to different trackers and enables flexible incremental expansion to any detector-tracker combination. We provide the detector-tracker used for each dataset in Table~\ref{tab:detail}. Among them, Refer-KITTI/v2~\cite{Wu_2023_CVPR,zhang2024bootstrappingreferringmultiobjecttracking} and LaMOT-MOT17~\cite{li2024lamotlanguageguidedmultiobjecttracking,milan2016mot16benchmarkmultiobjecttracking} redefine their test sets directly from the original training splits. Therefore, we retrain the corresponding detectors to avoid data leakage. All open-source and retrained tracking results are provided in the Supplementary Materials. The detector ``TempRMOT-D~\cite{zhang2024bootstrappingreferringmultiobjecttracking}" indicates its variant retrained on pure detection data, as we empirically find it performs well in autonomous driving scenarios. We further report FlexHook results on Refer-KITTI-V2~\cite{zhang2024bootstrappingreferringmultiobjecttracking} with a significantly weaker detector–tracker combinations (D-DETR~\cite{zhu2021deformabledetrdeformabletransformers} \& StrongSORT~\cite{du2023strongsort}). The comparison of pure tracking performance across different detector-tracker and its impact on FlexHook can be found in the Supplementary Materials.

\subsubsection{Implement Details} 

We set the input image resolution to $224\times672$, substantially reducing computational overhead compared to iKUN~\cite{Du_2024_CVPR}, which encodes $224\times224$ cropped patches and $672\times672$ global images. In our experiments, the trajectory segment length $p$ is set to 4, the number of sampled expressions $\hat{N}$ to 36, the number of reference points $M$ to 10, the expression length $L$ to 25. The $\lambda$, boundary $\delta$ and sharpness $\alpha$ in the regularization term $\mathcal{L}_r$ is 0.01, 0.1, and 30, respectively. The multi-scale feature maps consist of 4 levels, with the corresponding shape of the sampled target features $(h, w)$ set to $(16,48)$, $(8,24)$, $(4,12)$, and $(2,6)$. We adopt AdamW~\cite{loshchilov2019decoupledweightdecayregularization} as the optimizer with a learning rate of $3\text{e}{-5}$, and train the model for 20 epochs. All experiments are conducted on 2 NVIDIA RTX 4090 GPUs. We include pseudocode with full implementation details and hyperparameters in the Supplementary Materials.

\subsection{Benchmark Experiments}
\begin{table}[!ht]
\centering
\small
\caption{\textbf{Comparison with state-of-the-art methods} on Refer-Dance~\cite{Du_2024_CVPR} and LaMOT~\cite{li2024lamotlanguageguidedmultiobjecttracking}. ``$^\dagger$" denotes methods that are not task-specific, and \textbf{T} indicates ``Two-stage”.}
\begin{tabular}{c|c|cccc}
\toprule
Method & \textbf{T} & HOTA & DetA & AssA & LocA \\
\midrule
\multicolumn{6}{c}{Refer-Dance~\cite{Du_2024_CVPR}} \\

\midrule
TransRMOT~\cite{Wu_2023_CVPR}&\ding{56}&9.58&4.37&20.99&-\\
iKUN~\cite{Du_2024_CVPR}&\ding{51} &29.06&25.33&33.35&-\\
\midrule
\textbf{FlexHook-best}&\ding{51}&\textbf{32.17}&\textbf{28.39}&\textbf{36.52}&\textbf{91.31}\\
\midrule
\multicolumn{6}{c}{LaMOT~\cite{li2024lamotlanguageguidedmultiobjecttracking}} \\
\midrule
TransRMOT~\cite{zhang2024bootstrappingreferringmultiobjecttracking}&\ding{56}&27.74&39.56&21.33&80.17 \\

MOTRv2$^\dagger$~\cite{Zhang_2023_CVPR}&\ding{56}&45.19&49.67&41.19&89.86 \\

BYTETrack$^\dagger$~\cite{10.1007/978-3-031-20047-2_1}&\ding{51}&46.13&50.37&40.34&92.09\\
LaMOTer~\cite{Du_2024_CVPR}&\ding{51}&48.45&\textbf{53.50}&43.92&\textbf{92.91}\\
\midrule
\textbf{FlexHook-best}&\ding{51}&\textbf{56.77}&47.23&\textbf{68.24}&91.96\\
\bottomrule
\end{tabular} 

\label{tab:dance}
\end{table}

We compare FlexHook with existing task-specific models on Refer-KITTI/V2~\cite{zhang2024bootstrappingreferringmultiobjecttracking,Wu_2023_CVPR} in Table~\ref{tab:sota_refer_kitti_v2}. Whether using CLIP~\cite{pmlr-v139-radford21a} encoders or the combination of a RoBERTa~\cite{liu2019robertarobustlyoptimizedbert} text encoder with ResNet~\cite{He_2016_CVPR} and ROPE Swin-T~\cite{heo2024ropevit} visual backbones, FlexHook consistently achieves substantial performance gains over existing methods, thereby eliminating the strong dependency on CLIP. In particular, our best model, FlexHook with RoBERTa~\cite{liu2019robertarobustlyoptimizedbert} and ROPE Swin-T~\cite{heo2024ropevit} (FlexHook-best), maintains competitive performance even when paired with weaker detector–tracker settings (D-DETR~\cite{zhu2021deformabledetrdeformabletransformers} \& StrongSORT~\cite{du2023strongsort}). In non-autonomous-driving scenarios, we compare FlexHook-best with other methods on Refer-Dance~\cite{Du_2024_CVPR} and LaMOT~\cite{li2024lamotlanguageguidedmultiobjecttracking}, as shown in Table~\ref{tab:dance}. FlexHook outperforms the current state-of-the-art methods on both datasets, demonstrating its robustness in mixed-scenarios and large-scale data settings. The qualitative results can be found in the Supplementary Materials.

\subsection{Ablation Study}
\subsubsection{Main components}

\begin{table}[htbp]

    \small
    \centering
    \caption{Comparison with iKUN on Refer-KITTI-V2~\cite{zhang2024bootstrappingreferringmultiobjecttracking}. ``TI" indicates Temporal Integration, in which ``\ding{56}" denotes pooling.}
    \resizebox{\linewidth}{!}{
    \begin{tabular}{c|ccc|ccc}
    \toprule
    Method&C-Hook&PCD&TI&HOTA & DetA & AssA \\
    \midrule

    iKUN~\cite{Du_2024_CVPR}&\ding{56}&\ding{56}&\ding{56}&10.32&2.17&49.77\\
    \midrule
    \multirow{3}{*}{\textbf{FlexHook-iKUN}}&\ding{51}&\ding{56}&\ding{56}&34.49&22.51&52.97\\
    ~&\ding{51}&\ding{51}&\ding{56}&38.62&27.92&53.58\\
    ~&\ding{51}&\ding{51}&\ding{51}&\textbf{39.19}&\textbf{28.47}&\textbf{54.11}\\
    
    \bottomrule
    \end{tabular}}
    \label{tab:pd}
\end{table}

Table~\ref{tab:pd} evaluates the performance of the main components of FlexHook on Refer-KITTI-V2~\cite{zhang2024bootstrappingreferringmultiobjecttracking}. Here, FlexHook-iKUN is a variant specifically designed to emphasize C-Hook and PCD by comparing it with iKUN~\cite{Du_2024_CVPR}. It uses CLIP~\cite{pmlr-v139-radford21a} as the encoder, employs the same detector-tracker as iKUN~\cite{Du_2024_CVPR}, and disables Conditioning Enhancement. ``\ding{56}" in PCD column indicates that cosine similarity is used to compute the matching score instead of the PCD. The results show that C-Hook, PCD, and TI all bring significant performance improvements. Moreover, combined with FlexHook's performance under different encoder settings, it can be observed that PCD performs well not only in non-aligned spaces but also maintains strong performance in aligned embedding spaces, demonstrating its potential to replace cosine similarity as the core matching metric.

\begin{table*}[ht]
\centering
\centering
\small
\caption{The total time cost on Refer-KITTI-V2~\cite{zhang2024bootstrappingreferringmultiobjecttracking}. Since different paradigms involve distinct data formulation schemes (frame-level vs. bbox-level), we use the total training/inference time for clarity. FLOPs-based analysis can be found in the Supplementary Materials.}
\begin{tabular}{c|cc|c|c|ccc|c}
\toprule
Method &\makecell{Visual\\Backbone} &\makecell{Text\\Encoder} &Epoch&Train&Detect&Track& Refer&HOTA\\
\midrule
TempRMOT~\cite{zhang2024bootstrappingreferringmultiobjecttracking}&Trainable&Frozen&60&51.68 h&-&-& 256.61 min &35.04\\

TempRMOT-D~\cite{zhang2024bootstrappingreferringmultiobjecttracking}&Trainable&-&10&8.61 h& 2.72 min&-&-&-\\
iKUN~\cite{Du_2024_CVPR}$^\dagger$&Trainable&Partial&100& 2.46 h& 2.72 min&\textless 1 min& 136.70 min&10.32\\
\midrule
\multirow{2}{*}{\textbf{FlexHook-best}}&Trainable&Trainable&20&1.91 h& 2.72 min&\textless 1 min& \textbf{51.47 min}&\textbf{42.53}\\
~&Frozen&Frozen&20&\textbf{0.77 h}& 2.72 min&\textless 1 min& \textbf{51.47 min}&\underline{40.86}\\

\bottomrule
\end{tabular} 

\label{tab:time}
\end{table*}

\subsubsection{Neighboring and Conditioning Enhancement}
\label{sec:lr}
\begin{table}[!ht]
\small
    \centering
    \caption{Comparison of the performance about Neighboring and Conditioning Enhancement.}
    \begin{tabular}{c|c|cccc}
    \toprule
        Method & $M$ & HOTA & DetA & AssA  \\ \midrule
        \multirow{3}{*}{FlexHook-CLIP} & 0 & 40.59 & 27.87  &  \textbf{59.23} \\ 
        ~ & 10 & \textbf{41.42} &  \textbf{29.06} &  59.20 \\ 
        ~ & 30 & 40.82 & 28.78 & 58.01   \\ 
        \midrule
        \multirow{3}{*}{FlexHook-R 34} & 0 & 40.97 & 29.85 & 56.39  \\ 
        ~ & 10 &\textbf{41.50} &  \textbf{30.07} & 57.40  \\ 
        ~ & 30 &41.35 & 29.51 & \textbf{58.09}  \\ 
        \midrule
        \multirow{3}{*}{FlexHook-best} & 0 & 41.94 & 30.45 & 57.92   \\ 
        ~ & 10 & \textbf{42.53 }&  \textbf{30.63} & \textbf{59.19}  \\
        ~ & 30 &41.56 &  30.00 &  57.71 \\ 
        \midrule
        \makecell[c]{FlexHook-best\\w/o Neighbor} & 0 & 40.65 & 28.86 & 57.42 \\
        \bottomrule
    \end{tabular}
    \label{tab:hook}
\end{table}

We analyze the impact of Neighboring Grid Sampling and Conditioning Enhancement on Refer-KITTI-v2~\cite{zhang2024bootstrappingreferringmultiobjecttracking} in Table~\ref{tab:hook}, where $M$ denotes the number of sampling points used for Conditioning Enhancement, with $M=0$ indicating no enhancement. As observed, regardless of the vision backbone used, Conditioning Enhancement consistently yields robust improvements. This indicates that previous language-agnostic feature modeling strategies are insufficient for comprehensively handling complex expressions. The number of reference points has a relatively minor impact on model performance, and setting $M = 10$ is an empirical choice. In the setting without enhancement, removing the noise introduced in Neighboring Grid Sampling leads to a performance drop, highlighting the importance of simulating the noise inherent to the tracking task.

\subsubsection{Impact of Frozen Encoder}
\begin{table}[!ht]
\centering
\small
\caption{Performance of FlexHook after freezing different modules on Refer-KITTI-V2~\cite{zhang2024bootstrappingreferringmultiobjecttracking}.}
\resizebox{0.478\textwidth}{!}{\begin{tabular}{c|cc|ccc}
\toprule
Method&\makecell{Visual\\Backbone} &\makecell{Text\\Encoder}& HOTA & DetA & AssA \\
\midrule
\multirow{4}{*}{\makecell{FlexHook\\-CLIP}}&Trainable&Frozen&41.03&29.29&57.62\\
&Frozen&Trainable&41.38&29.21&58.76 \\
&Frozen&Frozen&40.95&\textbf{29.56}&56.85 \\
&Trainable&Trainable&\textbf{41.42}&29.06&\textbf{59.20} \\
\midrule
\multirow{4}{*}{\makecell{FlexHook\\-best}}&Trainable&Frozen&40.84&28.41&58.86\\
&Frozen&Trainable&41.83&30.02&58.41 \\
&Frozen&Frozen&40.86&28.16&59.44 \\
&Trainable&Trainable&\textbf{42.53}&\textbf{30.63}&\textbf{59.19} \\

\bottomrule
\end{tabular} }

\label{tab:freeze}
\end{table}



In Table~\ref{tab:freeze}, we evaluate the impact of freezing different encoders in FlexHook. Compared to CLIP~\cite{pmlr-v139-radford21a}, freezing the unaligned encoders causes a more significant performance drop. Meanwhile, freezing the text encoder has a greater negative impact than freezing the visual backbone, which can be attributed to the larger distribution gap in the linguistic space than that in the visual space between the pre-training datasets and the autonomous driving dataset. However, freezing any of the components results in only a modest performance drop, allowing a trade-off in performance for training efficiency.

\subsubsection{Training and Inference Time} Table~\ref{tab:time} shows a comparison of efficiency on our experimental machine. The detector–tracker combination used is TempRMOT-D~\cite{zhang2024bootstrappingreferringmultiobjecttracking} \& NeuralSORT~\cite{Du_2024_CVPR}. All experiments use the maximum batch size supported by the GPU to minimize the time cost. By eliminating the redundancy encoding stage and leveraging the unique advantage of PCD’s parallel processing, FlexHook achieves the fastest overall inference speed. During detector training, as the dataset contains only two ``expressions" (car and pedestrian), TempRMOT-D~\cite{zhang2024bootstrappingreferringmultiobjecttracking} can converge in fewer than 10 epochs, instead of the 60 epochs used in its RMOT version. The speed is generally faster with other detectors such as Deformable-DETR~\cite{zhu2021deformabledetrdeformabletransformers}. Considering that most TBD trackers are training-free, even when built from scratch, FlexHook still achieves the lowest total training time, which becomes more pronounced in incremental deployment scenarios. When freezing encoders, FlexHook allows for further improvement in efficiency with only a minor performance sacrifice.

\section{Conclusion}


In this work, we address the limitations of existing two-stage RBT methods in RMOT by proposing FlexHook, a flexible hook-like framework. FlexHook introduces two components, C-Hook and PCD, which revolutionize the feature construction and correspondence decoding pipelines, achieving significant gains in both accuracy and computational efficiency. By fully leveraging the capabilities of pre-trained models, injecting textual prior information, and reducing dependence on CLIP~\cite{pmlr-v139-radford21a}, FlexHook demonstrates excellent performance across multiple benchmarks and makes two-stage RBT strong again.

{
    \small
    \bibliographystyle{ieeenat_fullname}
    \bibliography{main}
}


\end{document}